\documentclass[10pt,twocolumn,letterpaper]{article}

\usepackage[pagenumbers]{cvpr} 

\usepackage{graphicx}
\usepackage{amsmath}
\usepackage{amssymb}
\usepackage{booktabs}
\usepackage{url}

\usepackage[pagebackref,breaklinks,colorlinks]{hyperref}

\usepackage[capitalize]{cleveref}
\crefname{section}{Sec.}{Secs.}
\Crefname{section}{Section}{Sections}
\Crefname{table}{Table}{Tables}
\crefname{table}{Tab.}{Tabs.}

\def\cvprPaperID{4049} 
\def\confName{CVPR}
\def\confYear{2022}

\begin{document}

\title{Structure-Aware Flow Generation for Human Body Reshaping}

\author{Jianqiang Ren,~~Yuan Yao,~~Biwen Lei,~~Miaomiao Cui,~~Xuansong Xie \\
DAMO Academy,~~Alibaba Group\\
{\tt\small \{jianqiang.rjq, ryan.yy, biwen.lbw, miaomiao.cmm\}@alibaba-inc.com, xingtong.xxs@taobao.com}
}
\maketitle

\begin{abstract}
Body reshaping is an important procedure in portrait photo retouching. Due to the complicated structure and multifarious appearance of human bodies, existing methods either fall back on the 3D domain via body morphable model or resort to keypoint-based image deformation, leading to inefficiency and unsatisfied visual quality. In this paper, we address these limitations by formulating an end-to-end flow generation architecture under the guidance of body structural priors, including skeletons and Part Affinity Fields, and achieve unprecedentedly controllable performance under arbitrary poses and garments. A compositional attention mechanism is introduced for capturing both visual perceptual correlations and structural associations of the human body to reinforce the manipulation consistency among related parts. For a comprehensive evaluation, we construct the first large-scale body reshaping dataset, namely BR-5K, which contains 5,000 portrait photos as well as professionally retouched targets. Extensive experiments demonstrate that our approach significantly outperforms existing state-of-the-art methods in terms of visual performance, controllability, and efficiency. The dataset is available at our website: \url{https://github.com/JianqiangRen/FlowBasedBodyReshaping}.
\end{abstract}

\begin{figure}[t]
  \centering
  \resizebox{0.98\linewidth}{!}{
   \includegraphics{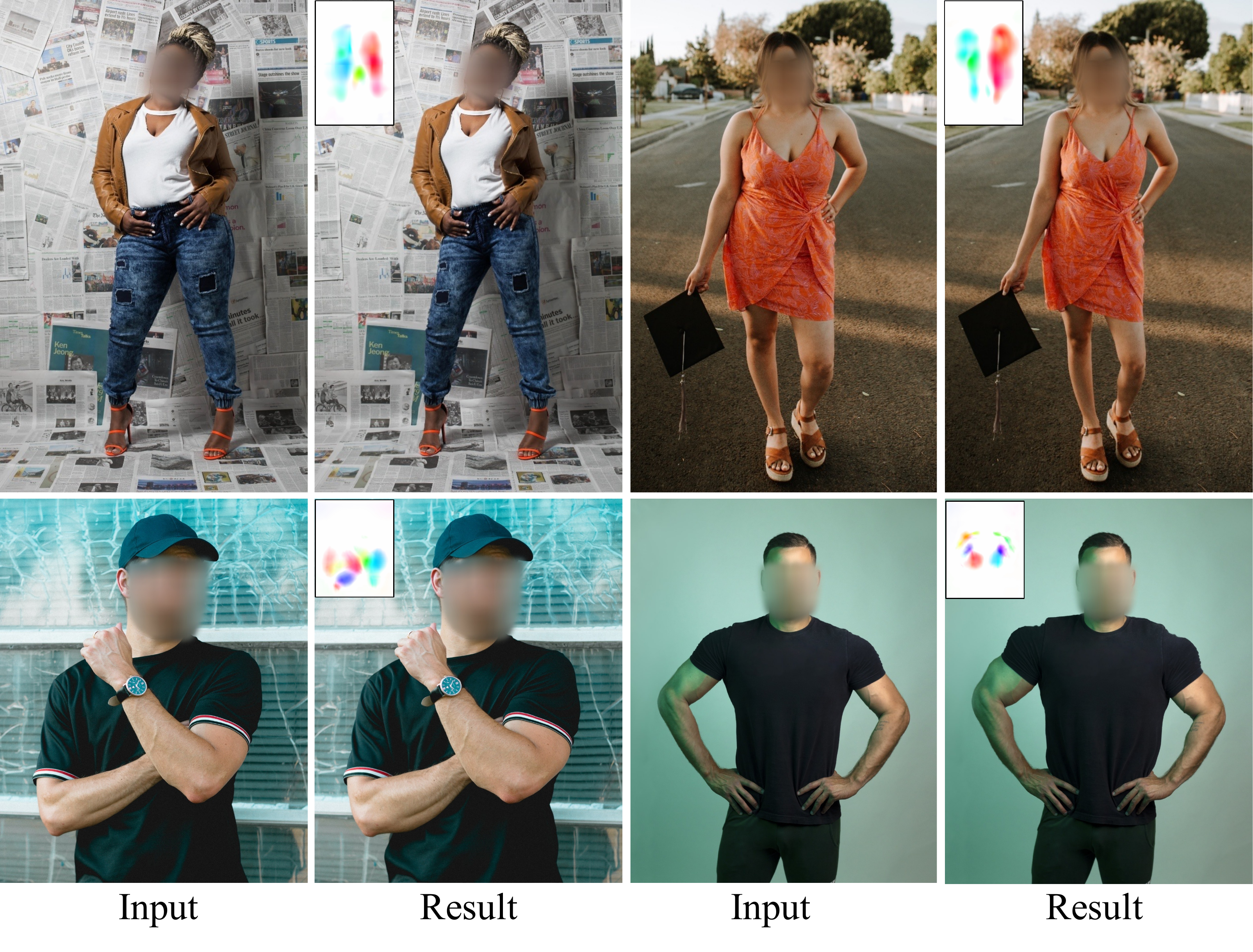} }
   \vspace{-10pt}
  \caption{Our approach provides an efficient and controllable solution for human body reshaping, allowing figures to be more slender (\emph{top}) or muscular (\emph{bottom}). The corresponding deformation flows are showcased in the top-left corners. All of the images presented in this paper are included in the proposed BR-5K dataset.}
  \label{fig: showCases}
  \vspace{-15pt}
\end{figure}

  \vspace{-15pt}
\section{Introduction}
\label{sec:intro}

Portrait photography retouching is widely used in various fields, such as social media, weddings, and advertisements. While automatic face reshaping is well researched\cite{xiao2020deep,wang2019detecting,yi2020animating,shih2019distortion} and employed in many applications (\emph{e.g.}, Instagram, Tictok), automatic body reshaping, which generally aims to make figure looks more shapely and attractive, is still far from being settled due to the complicated structure and appearance of human bodies. In fact, body reshaping is a troublesome task even for professional artists. The artists usually rely on Liquify tool in image editing software, adjust the distortion brush frequently, and manually edit through every single part of body areas. It takes considerable skill and time to achieve a visually pleasing result. 

Traditional solutions for body reshaping task can be categorized into either rule-based or 3D model-based methods. The rule-based methods employ non-rigid image deformation\cite{ma2013nonrigid,schaefer2006image} according to a set of body contour keypoints following some hand crafted rules, and 3D model-based approaches\cite{zhou2010parametric,yang2014semantic,richter2012real,xu2019human} try to reconstruct 3D morphable model from a single portrait image, and adjust body shape by manipulating a small set of parameters corresponding to body semantic attributes. Unfortunately, neither of them can satisfy the quality and efficiency requirements in practice. Rule-based methods can only work under standard poses and garments, and body contour detection is not robust enough for generating a reliable, quality-assured reshaping result. 3D model-based methods are learned from a limited number of full-body scans so that cannot span the entire human shape space, especially for extreme camera directions and poses. Moreover, precisely recovering and registering a 3D model to image is challenging as well, generally needing considerable user assistance for pose and shape fitting\cite{zhou2010parametric}, or extra depth information\cite{xu2019human,richter2012real}.

Powered by deep generative technology\cite{isola2017image, zhu2017unpaired, wang2018high}, several methods\cite{zhu2019progressive, ma2017pose} propose to directly generate person images guided by pose information. Nevertheless, they are still struggling for synthesizing photorealistic person images to avoid visible misalignment artifacts. Recent methods\cite{ren2020deep,tang2021structure,wang2019detecting,yi2020animating} have revealed the effectiveness of flow field in spatial deformation tasks. However, feature-level flow-based methods\cite{ren2020deep,tang2021structure} can only afford to generate low-resolution person images, thus hardly satisfying the body retouching demands in practice. While pixel-level flow-based models\cite{wang2019detecting,yi2020animating} can conduct high-resolution face manipulation, their performance on body is far from our expectation, resulting in unsatisfied or inconsistent reshaping results as we observed. We speculate the reason is that the regular physiological structure of the face dramatically reduces the difficulty to learn face editing, but the body is much more complicated due to the flexible articulated structure and multifarious garments.

To address the aforementioned challenges, we propose a structure-aware flow generation framework for body reshaping. Given a portrait photo, we extract body skeletons and Part Affinity Fields (PAFs) as important structural priors to guide flow generation. To encourage deformation consistency among body parts, a \emph{Structure Affinity Self-Attention (SASA)} mechanism is introduced to capture long-range perceptual correlations and structural associations simultaneously. Finally, with the predicted flow field, which is both globally coherent and locally smooth, we generate the reshaped body using a warping operation. Moreover, our method can efficiently handle with high-resolution (4K) images by predicting flows on small images and applying the up-sampled flows to the original high-resolution images.

To facilitate the research on this important task, we create the first large-scale body retouching dataset BR-5K, which comprises 5,000 individual portrait photos as well as their retouched targets edited by professional artists. Extensive experiments demonstrate the effectiveness of the proposed method and dataset. The contributions of this paper are summarized as follows:
\begin{itemize}
\item We present the first 2D end-to-end structure-aware flow generation framework for body reshaping, which significantly exceeds state-of-the-art methods in terms of visual quality, controllability, and efficiency.

\item We demonstrate that utilizing the skeletons and PAFs as structural priors can significantly alleviate the difficulty for body manipulation learning, leading to more accurate pixel-level flow prediction. A compositional attention mechanism \emph{SASA}, which considers both perceptual correlations and structural associations, is further proposed to reinforce the manipulation consistency across related human parts.

\item We create the first large-scale body retouching dataset BR-5K,  which contains 5,000 high-resolution individual portrait photos as well as their retouched targets edited by professional artists.
\end{itemize}

\section{Related Works}

\noindent {\bf Portrait Reshaping.} Reshaping human portraits has been widely used in digital entertainment, and photography production. By using image editing software such as Adobe Photoshop, it often involves a set of semantic-aware local modifications, demanding professional skills. Many works have been proposed to tackle the complex issue by first focusing on the face reshaping task. Leyvand~\etal~\cite{leyvand2008data} propose a data-driven facial beautification method by deforming the facial shape of an input image using a landmark-based 2D warping function. SHIH~\etal~\cite{shih2019distortion} introduce a content-aware warping method combining the stereographic and perspective projections onto a single image to correct the wide-angle distortion of faces. However, methods operate directly on the 2D image data are limited in frontal views or accurate face segmentations. Liao~\etal~\cite{liao2012enhancing} extend the geometric modification onto 3D face model, providing the possibility to handle non-frontal views. Zhao~\etal~\cite{zhao2018parametric} present to manipulate the deformed 3D face by utilizing a soft tissue-based regression model defined on a sparse set of facial landmarks. Xiao~\etal~\cite{xiao2020deep} leverage the parametric face representation based on 3DMM~\cite{blanz1999morphable} and the facial expression model to empower dense control of face shape. Tang~\etal~\cite{tang2021parametric} further address the consistency and coherency problem in video portrait reshaping task. Compared with face reshaping, body shape editing has been much less explored. Zhou~\etal~\cite{zhou2010parametric} propose a body-aware image warping method to create desired reshaping effects by integrating a 3D whole-body morphable model. Our method directly manipulates the 2D body image via an end-to-end warping pipeline.

\begin{figure*}[ht]
  \centering
  \resizebox{0.98\linewidth}{!}{
   \includegraphics{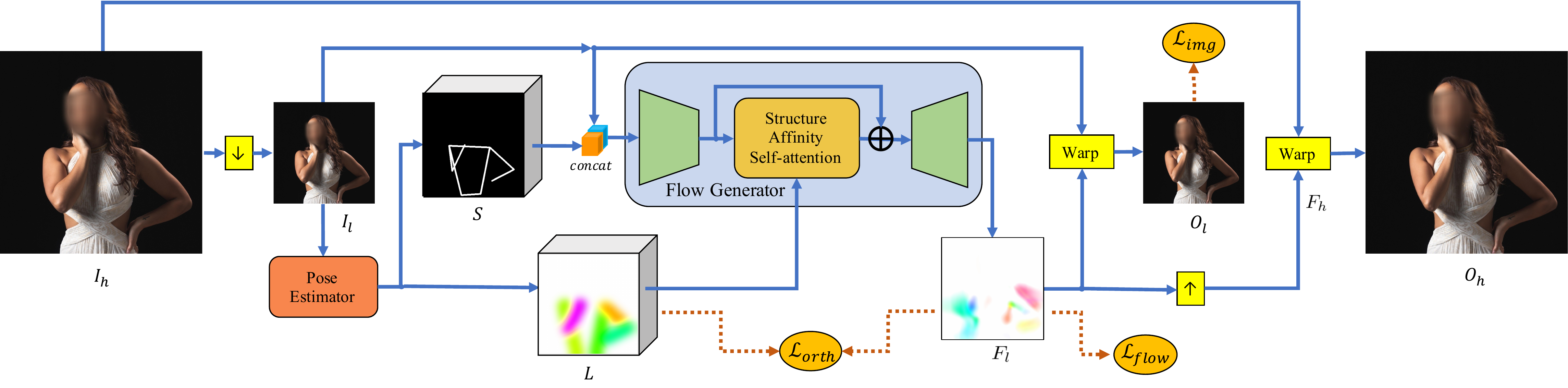} }
  \caption{Overview of the proposed method. Given a high-resolution portrait image $I_{h}$, we first extract its skeleton maps $S$ and PAFs $L$ after down-sampling, and then generate deformation flow $F_{l}$ by feeding the concatenation of $I_{l}$ and $S$ into the Flow Generator. A compositional attention module (SASA) that combines perceptual correlation and structural association is introduced in the bottleneck layer to enhance manipulation consistency among related body parts. Finally, we upsample the low-resolution flow $F_{l}$ to $F_{h}$ and conduct a warping operation to $I_{h}$ to obtain the final result. The orange dotted line denotes the flow of loss functions.}
  \label{fig: architecture}
  \vspace{-15pt}
\end{figure*} 

\noindent {\bf Flow-based Image Transformation.} A closely related task with body shape transform is pose guided person image generation task, whose goal is to transfer a person image from the source pose to a new target pose. Motivated by the development of GANs~\cite{goodfellow2020generative}, many researchers attempted to tackle the problem following the image-to-image transformation paradigm~\cite{isola2017image,zhu2017unpaired,wang2018high}. However, these generation-based models~\cite{ma2017pose,zhu2019progressive} may fail to handle the feature misalignment caused by the spatial deformation between persons with different poses. Flow-based methods have recently been proposed to leverage appearance flow to solve the spatial transformation problem. Ren~\etal~\cite{ren2020deep} propose a global-flow local-attention framework to reassemble the source feature patch to the target. Tang~\etal~\cite{tang2021structure} design a structure-aware framework and decompose the task of learning the overall appearance flow field into learning different local flow fields for different semantic body parts. Although sharing similar dependencies such as pose structure, flow-based person pose transforms utilize pose information as input condition and warp image at feature level, which may lead to blurry results lacking vivid appearance details and limited in handling low-resolution images. Our proposed method uses pose information as priors for consistency enhancement and warps image at the pixel level, capable of tackling high-resolution images.

Our method also has ties to ATW~\cite{yi2020animating} and FAL~\cite{wang2019detecting}. ATW proposes a ResWarp module to warp residuals based on the predicted flow motion and adds the warped residuals to generate the final HD results, which enables efficient face animation of high-resolution images. FAL is present to detect image warping manipulation applied to faces with Photoshop. It uses a local warping field prediction network to localize and undo face manipulations. Compared with face transformation using optical flow, body transformation involves rigour challenges for its complex structures and correlations between body parts.

\noindent {\bf Attention Mechanism.} With its effective learning capability, attention mechanism has been widely employed in various computer vision tasks~\cite{xu2015show,yang2016stacked,zhao2019pyramid,yu2018generative}. One way that incorporates attention mechanism is to use spatial attention that reweights the convolution activations\cite{wang2018non,zhang2019self}, which provides a possibility for long-range modeling across distant regions. The other way is to model interdependencies between channels\cite{hu2018squeeze,hu2018gather}. There are other works that combine the two ideas~\cite{chen2017sca,fu2019dual,woo2018cbam}. Our method adopts a similar framework from CoDA~\cite{tay2019compositional}, which proposes a compositional quasi-attention mechanism with a dual affinity scheme.

\section{Datasets}
We build BR-5K, the first large-scale dataset for body reshaping. It consists of 5,000 high-quality individual portrait photos at 2K resolution collected from Unsplash\cite{Unsplash}. Since face is irrelevant to our task, we conduct face obfuscation for privacy protection. As body retouching is mainly desired by females, the majority of our collection are female photos, considering the diversity of ages, races (African:Asian:Caucasian = 0.33:0.35:0.32), poses, and garments. We invite three professional artists to retouch bodies using Photoshop independently, with the goal of achieving slender figures that meet the popular aesthetics, and select the best one as ground-truth. More details can be found in the supplemental file.

It is worth noting that the so-called ground-truth is not a standard for body shape, but rather a specific retouching style for evaluating the learning ability of reshaping models. In addition, although there is only one ground-truth for each image during training, our model can achieve continuous body adjustment to cater to various aesthetics.

\section{Methods}

The overall network architecture of our approach is illustrated in Figure~\ref{fig: architecture}. Given a high-resolution input image $I_{h}$, we first downsample it to $I_{l}$ with a lower resolution for computation efficiency. A redesigned \emph{pose estimator} module based on \cite{cao2017realtime} is proposed to extract the skeleton maps $\textbf{S}$ and the Part Affinity Fields (PAFs) $\textbf{L}$ from $I_{l}$, as we discover that skeletons are good at demarcating local warping orientations and that PAFs are helpful to locating where should be manipulated. Then we concatenate the recombined skeleton maps $\textbf{S}$ with $I_{l}$ and feed them into the Flow Generator. A \emph{Structure Affinity Self-Attention} module is introduced in the bottleneck of Flow Generator to boost the consistency of generated flow field $F_{l}$ under the guidance of the PAFs. Once the flow field $F_{l}$ is generated, we upsample it to the original size of $I_{h}$ and conduct a warping operation $\mathcal{W}(I_{h}; F_{h})$ to get the final retouched result $O_h$.

 \begin{figure}[t]
  \centering
  \resizebox{0.98\linewidth}{!}{
   \includegraphics{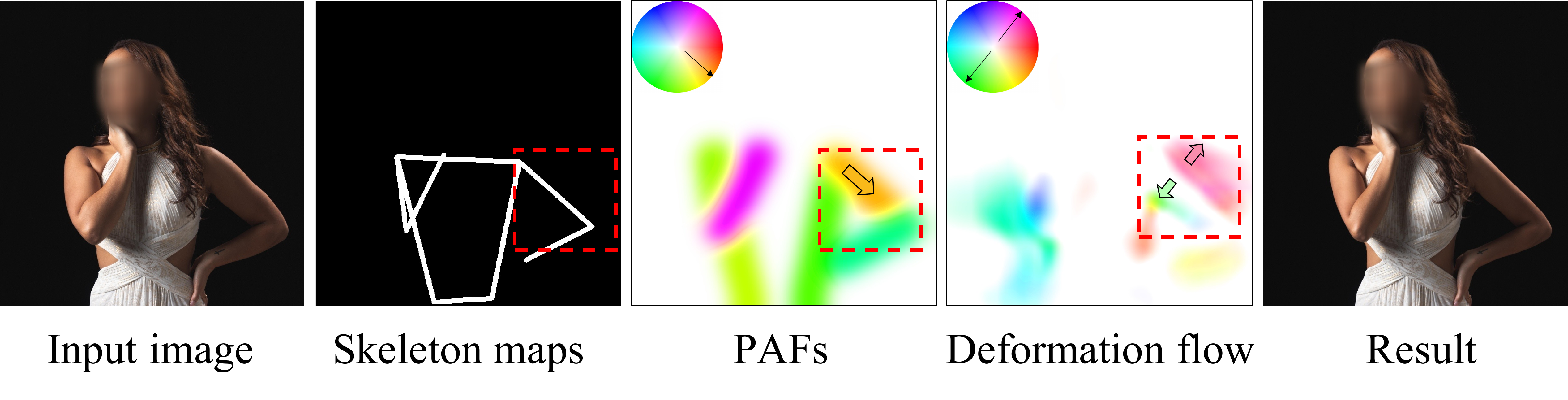} }
  \caption{The skeletons work as low-level features that are able to indicate opposite boundaries of deformation flow, while PAFs help to highlight where should be manipulated, and have orthogonality relation with the desired deformation flow (see the arrows in dashed red boxes and flow color coding circles).  }
  \label{fig: pose_prior}
  \vspace{-10pt}
\end{figure} 

\subsection{Pose Estimator}
Human 2D pose estimation is generally represented by a group of anatomical keypoints corresponding to different body parts. To better learn the associations between parts, Cao \etal~\cite{cao2017realtime} present Part Affinity Fields, a set of 2D vector fields that encode location and orientation of human limbs, which play an important role in their bipartite matching procedure. Similar to \cite{cao2017realtime}, we design the pose estimator as a two-branch convolutional network which allows predicting keypoint heatmaps and PAFs of the down-sampled portrait image $I_l$ simultaneously, and then we take both keypoints and PAFs as structural priors to guide body manipulation.

To further integrate discrete keypoints into skeletons, we predefine $N_{s}$ skeletons based on body anatomical structure and generate a stack of skeleton maps $\textbf{S}=({S}_{1},{S}_{2},...,{S}_{N_{S}})$ by linking keypoints belonging to the same limb, except for the keypoints on the head, and setting the values of the background pixels as zeros. Each channel of $\textbf{S}$ represents a skeletal bone of the body. As shown in Figure~\ref{fig: pose_prior}, we find that these skeletons are of great benefit to locating the local orientation boundaries of deformation flow, as the flow directions are generally contrary to each other on the opposite sides. We thus hypothesize that the skeletons provide low-level clues which help to indicate flow edges.

Different from skeletons, PAFs manifest more structural characteristics at the regional level. We modify and eliminate some fields from original PAFs which are irrelevant to our task (\emph{e.g.}, fields on head) and reassemble the others as our PAFs $\textbf{L}= ({L}_{1},{L}_{2},...,{L}_{N_{L}})$. Each channel of $\textbf{L}$ is a 2D vector field corresponding to a limb and additionally dilated 3 times to ensure that the limb is wholly included in it.
 
We convert the PAFs from Cartesian coordinate into Polar form, and fully utilize their magnitude fields and orientation fields respectively. As illustrated in Figure~\ref{fig: pose_prior}, we find that the magnitude of PAFs helps to highlight where should be manipulated in the image, and the orientation of PAFs is roughly perpendicular to the deformation flow around body areas. Therefore, we introduce the magnitude of PAFs to our attention mechanism while employing the orientation as an orthogonality constraint subsequently.

\begin{figure}[t]
  \centering
  \resizebox{0.98\linewidth}{!}{
   \includegraphics{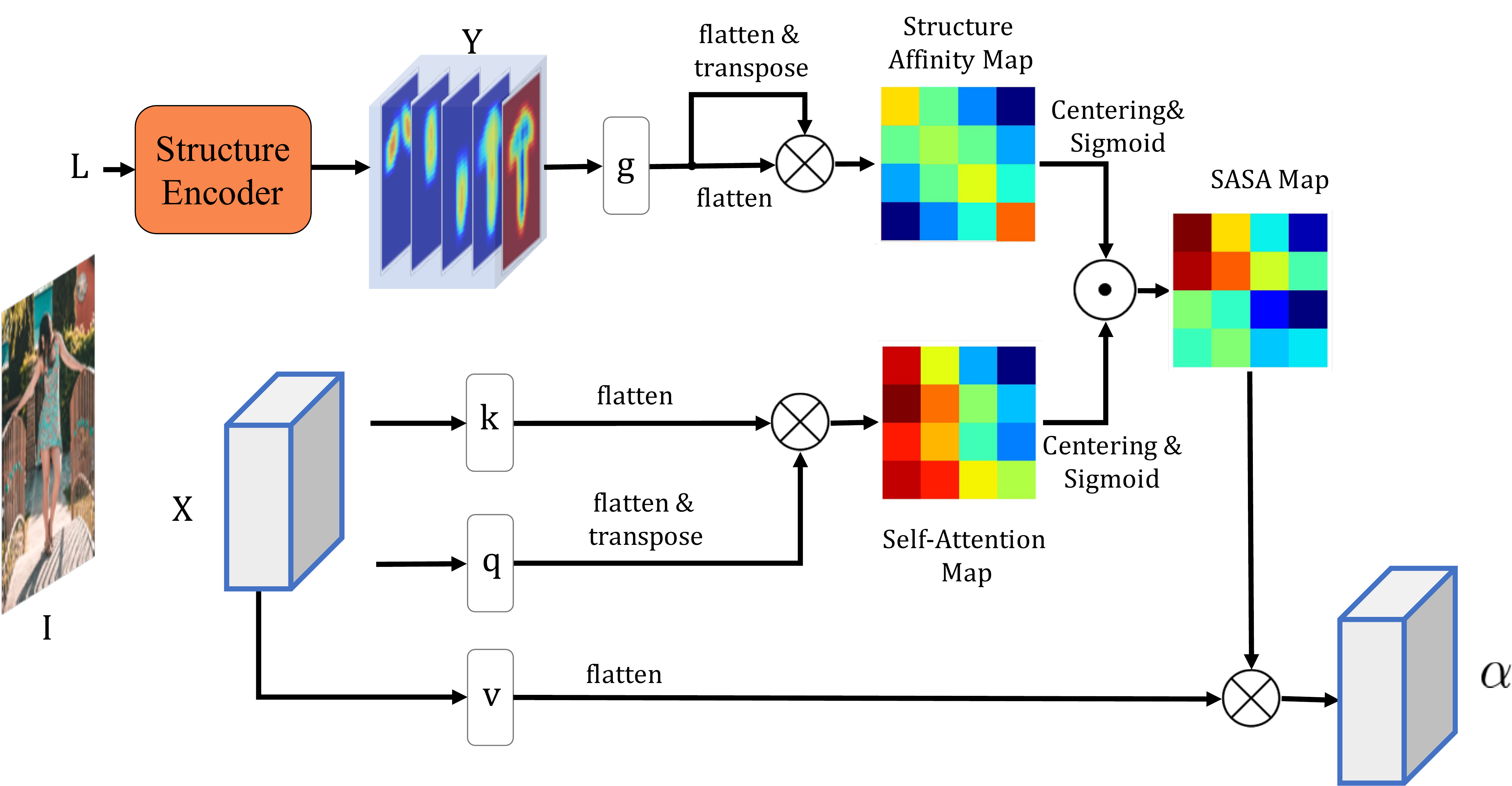} }
  \caption{The architecture of the \emph{SASA} module. With the structure heatmaps $Y$ and the deep feature maps $X$ as inputs, \emph{SASA} computes the attention residual $\alpha$ by a multiplicative composition of \emph{Structure Affinity Map} and \emph{Self Attention Map}. The $\otimes$ and the $\odot$ denote matrix and element-wise multiplication respectively.}
  \label{fig: sasa}
  \vspace{-10pt}
\end{figure}

\subsection{Structure Affinity Self-Attention}
Once the skeleton maps $\textbf{S}$ and PAFs $\textbf{L}$ are obtained, we concatenate the $\textbf{S}$ with $I_l$, and feed them into an encoder-decoder network to predict deformation flow, with the help of high-level structural priors provided by PAFs $\textbf{L}$. 

To reshape the human body properly, we wish the generated flow to be consistent among body parts. Specifically, the consistency means the manipulation should be smooth in each limb (\emph{i.e.}, no discontinuous artifacts) and be coherent across limbs (\emph{i.e.}, avoiding irrationally thick arms along with thin legs). We employ a compositional attention\cite{tay2019compositional} mechanism for this purpose. Self-attention\cite{wang2018non,zhang2019self} is initially designed for capturing long-range dependencies. However, it learns to allocate attention mainly according to the similarity of color and texture as observed in \cite{zhang2019self}, neglecting spatial structural associations. To further integrate structural priors with non-local attention, we present the \emph{Structure Affinity Self-Attention (SASA)} module inspired by CoDA\cite{tay2019compositional}, as depicted in Figure~\ref{fig: sasa}. By leveraging a compositional quasi-attention mechanism that composes the self-attention map with a structure affinity map, \emph{SASA} takes both visual perception and structural association of the input image into account. 

Let $X=Enc(concat(I_l,\textbf{S}))\in\mathbb{R}^{H\times{W}\times{C}}$ represents the implicit feature maps extracted by encoder, the \emph{Self Attention Map} $\Phi_{att} \in\mathbb{R}^{HW\times{HW}}$ is calculated as:

\begin{equation} \label{eq: att}
\Phi_{att} = \theta_k(X) \ast \theta_q(X)^T\,
\end{equation}
where $\theta$ denotes the convolution and flatten operation along the channel
dimension, $k$ and $q$ are $1\times1$ convolution kernels, $T$ indicates transpose operation. $\Phi_{att}$ reveals the correlation between image regions in visual perception aspect.

With regard to structural association, we calculate structure heatmaps $Y$ from PAFs $\textbf{L}$ by a pre-defined body \emph{structure encoding} (corresponding to \emph{Structure Encoder} block in Figure~\ref{fig: sasa}). Specifically, we get each limb's mask $M$ according to its PAF magnitude, and produce the structure heatmaps $Y$ by stacking the fore/back-ground mask with integrated masks of closely related body parts (left and right arms, torso, left and right leg).
 
\begin{equation} \label{eq: heatmaps}
Y = ({M}_{arms},{M}_{torso}, {M}_{legs}, {M}_{FG},{M}_{BG})\,
\end{equation}

We restrict $Y$ to keep the same $H$ and $W$ with $X$. Each heatmap in $Y$ highlights a union spatial distribution of related structure components. We define the \emph{Structure Affinity Map} $\Phi_{aff} \in\mathbb{R}^{HW\times{HW}}$ as following:

\begin{equation} \label{eq: aff}
\Phi_{aff} = \theta_g(Y) \ast \theta_g(Y)^T\,
\end{equation}
where $g$ is a $1\times1$ convolution kernel as well. The $\Phi_{aff}$ measures the structural associations between regions, where a higher affinity value indicates a stronger structural relationship. We then generate the \emph{SASA Map} $\Phi_{sasa}$ via a multiplicative composition following CoDA\cite{tay2019compositional}: 
\begin{equation} \label{eq: composition}
\Phi_{sasa} = sigmoid(\hat{\Phi}_{att}) \odot sigmoid(\hat{\Phi}_{aff})\,
\end{equation}
where $\odot$ denotes the element-wise multiplication operator, and $\hat{\Phi}_{\{aff, att\}}$ are calculated by a centering operation:

\begin{equation} \label{eq: centering}
\hat{\Phi}_{\{aff, att\}} = \Phi_{\{aff, att\}} - mean(\Phi_{\{aff, att\}})\
\end{equation}

\begin{figure}[t]
  \centering
  \resizebox{0.98\linewidth}{!}{
   \includegraphics{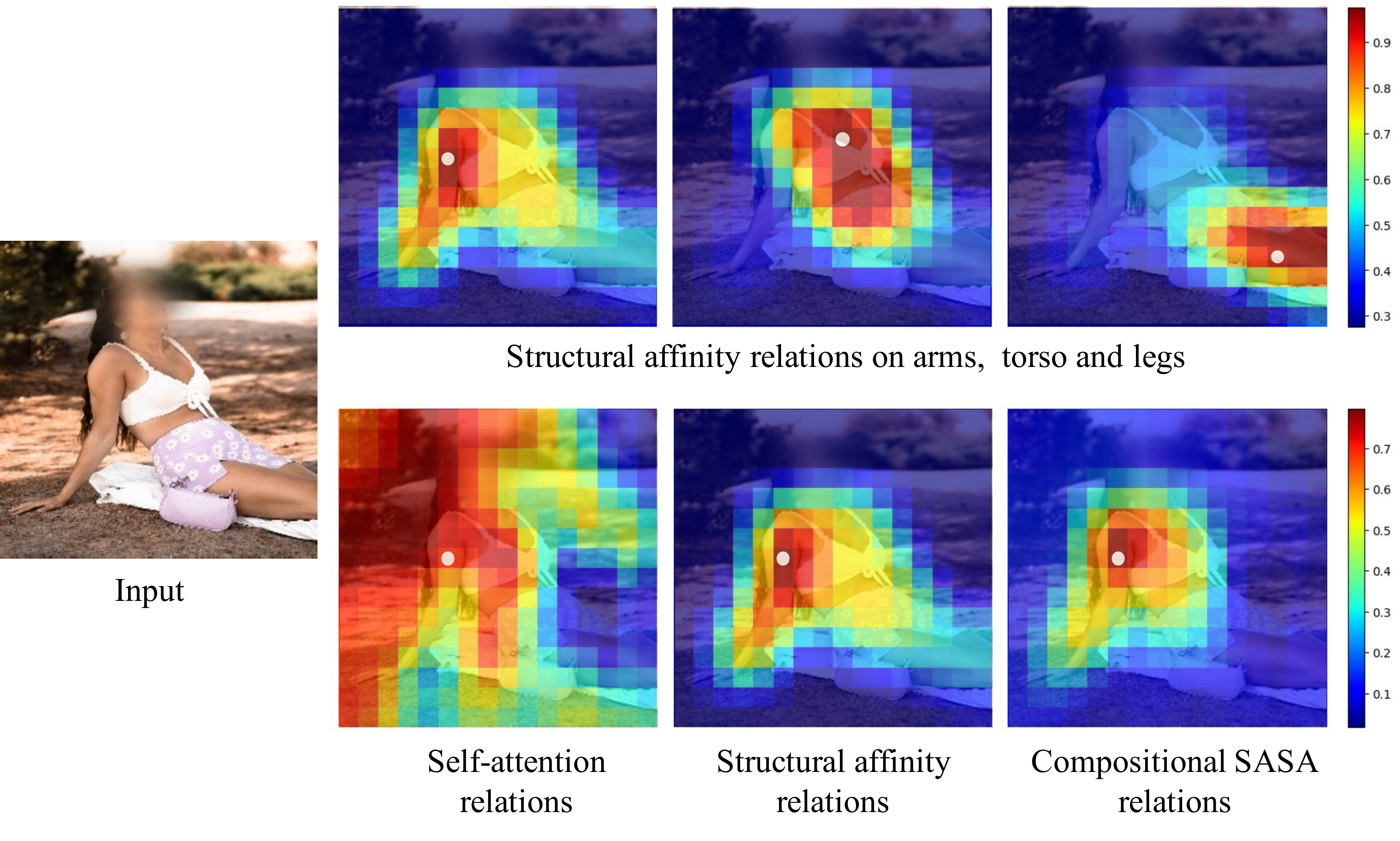} }
       \vspace{-5pt}
  \caption{The \emph{structure affinity} tends to allocate attention to related body parts, and self-attention works mainly according to perceptual similarity. The composition of them helps to erase some irrelevant attention relations.}
  \label{fig: attention_vis}
  \vspace{-15pt}
\end{figure}

The value of \emph{SASA} output $\alpha$ at the $i^{th}$ location can be calculated:

\begin{equation} \label{eq: alpha}
\alpha_{i} = \sum_{j=1}^{HW} \Phi_{sasa}(i,j) \ast \theta_{v}(X)_{j}\
\end{equation}
Finally, we add the scaled $\alpha$ back to the input $X$ to get the output feature maps:

\begin{equation} \label{eq: add_residual}
\hat{X} = X + \gamma \alpha,
\end{equation}
where $\gamma$ is a learnable scalar as \cite{zhang2019self}. Figure~\ref{fig: attention_vis} shows the visualization of attention relation maps of \emph{SASA}. We notice that the \emph{Structure Affinity} tends to allocate attention weights on structural-related regions of the body, and the Self-Attention works mainly according to the similarity of color or texture. The composition of them helps to erase some irrelevant attention relation between body and background caused by their visual similarity, and make the attention focus on closely related body areas, leading to accurate and consistent reshaping results, as shown in Figure~\ref{fig: ablations}.

\subsection{Flow Generation and Warping}
Once the feature maps $\hat{X}$ are obtained, we feed them into the decoder to generate the deformation flow field $F_{l}$, which has the same size with $I_{l}$. By conducting a warping operation $\mathcal{W}$, we can obtain the low-resolution reshaped output $O_{l}$:

\begin{equation} \label{eq: warp}
O_{l} =  \mathcal{W}(I_{l}; \mu F_{l})\
\end{equation}
where $\mu$ is an additional multiplier. The absolute value of $\mu$ can adjust flow magnitude while the sign of $\mu$ determines the flow direction. Intuitively, a larger $|\mu|$ will lead to a bigger change on body shape, and the positive/negative sign controls whether to lose or gain weight. The $\mu$ is set to 1.0 during training to imitate manually-retouched strategies, while it provides continuous controllability when editing a new portrait photo in practical applications (Figure~\ref{fig: runtime_control}).

Instead of directly upsampling the image $O_{l}$ to generate the high-resolution result, we upsample the flow $F_{l}$ to $F_{h}$ using bilinear interpolation and apply another warping operation to the source input $I_h$ to keep its original details:

\begin{equation} \label{eq: warp_h}
O_{h} =  \mathcal{W}(I_{h}; \mu F_{h})\
\end{equation}

Although the details of images are hard to recover via commonly upsampling, we observed the upsampled deformation flows are good enough to achieve compelling reshaping results on even 4K-resolution images, owing to the local smoothness of the generated flows, which is reasonable for avoiding discontinuous deformation artifacts. Visual examples of 4K-resolution image manipulation are provided in the supplementary material.

\subsection{Learning Strategies}

To encourage a high fidelity retouching output near to manual manipulation, we try to minimize the L1 distance between the low-resolution output $O_l$ and its corresponding ground-truth $\hat{O_l}$ (downsampled from high-resolution ground-truth $\hat{O_h}$):

\begin{equation} \label{eq: reconstruction}
\mathcal{L}_{img} = ||\hat{O_l} - O_l||_{1}\,
\end{equation}

We estimate the optical flow between the original and manually-retouched image using PWC-Net\cite{sun2018pwc}, and treat the estimated flow as ground-truth deformation flow $\hat{F_l}$ for providing the Flow Generator with more direct guidance. The objective function is:

\begin{equation} \label{eq: flow_loss}
\mathcal{L}_{flow} = ||\hat{F_l} - F_l||_{1}\,
\end{equation}

Since our method will adjust each limb's width but keep length and direction unchanged, the orientation of deformation flow and corresponding limb tend to be perpendicular to each other. Considering the PAFs represent the orientation of human limbs, we calculate the cosine similarity between PAFs and $F_l$, and encourage the orthogonality of them:

\begin{equation} \label{eq: flow_orth}
\mathcal{L}_{orth} = \frac{(\sum_{n=1}^{N_{L}} \textbf{L}_{n}) \cdot F_l}{|\sum_{n=1}^{N_{L}} \textbf{L}_{n}|\cdot |F_l|}\,
\end{equation}

The total loss function are given by:
\begin{equation} \label{eq: overall_loss}
\mathcal{L} = \lambda_{img}\mathcal{L}_{img}  + \lambda_{flow}\mathcal{L}_{flow} + \lambda_{orth}\mathcal{L}_{orth}\,
\end{equation}
where $\lambda_{img}$, $\lambda_{flow}$ and $\lambda_{orth}$ are the balancing parameters.

\begin{figure*}[ht]
  \centering
  \resizebox{0.98\linewidth}{!}{
   \includegraphics{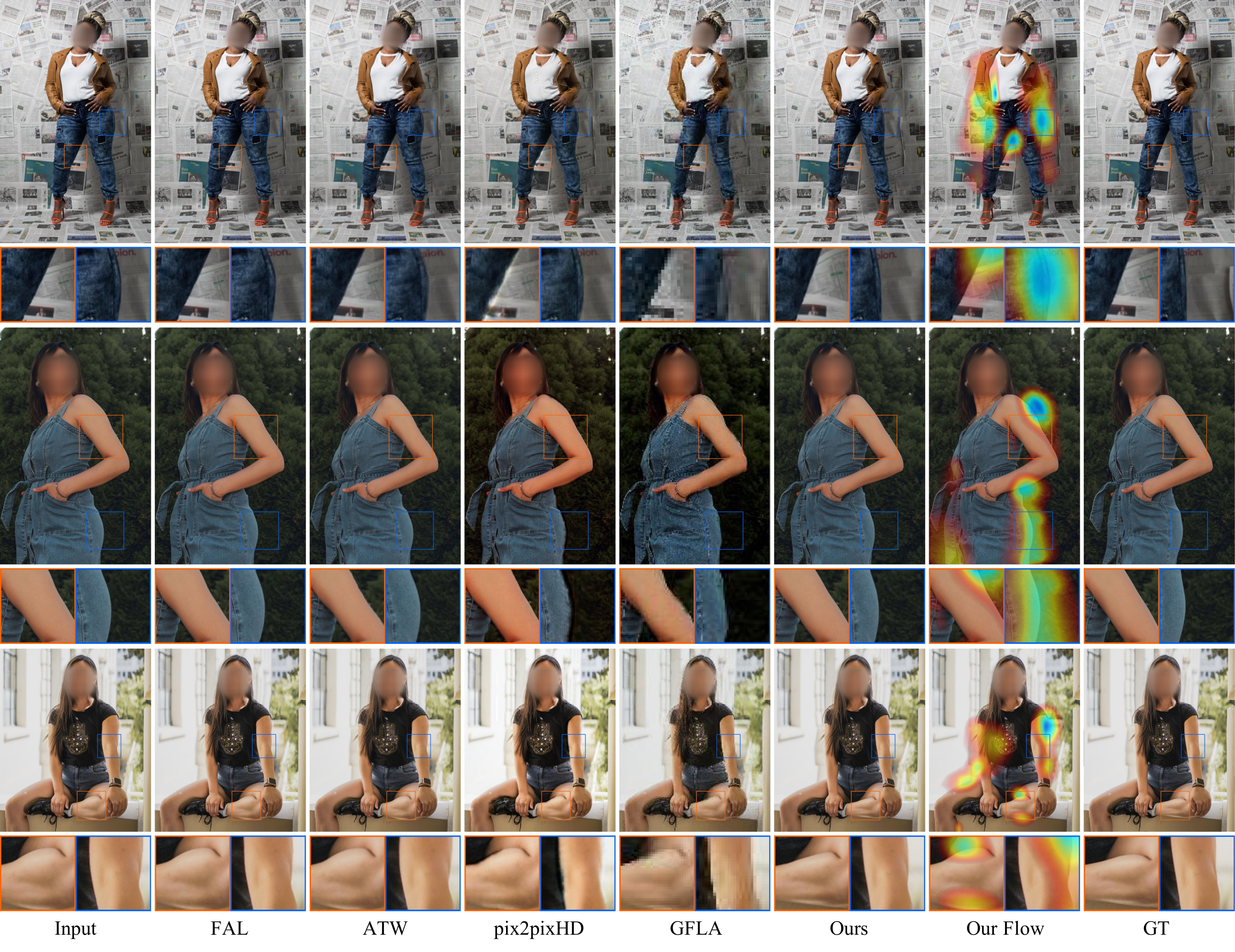} }
     \vspace{-10pt}
  \caption{Visual comparisons among different methods. Our method can produce high-resolution, believable, and consistent body reshaping results. The `Our Flow' column illustrates where are edited by our method. Zoom in for details.}
  \label{fig: cmp}
  \vspace{-15pt}
\end{figure*} 

\section{Experiments}
\subsection{Datasets and Settings}
\noindent {\bf BR-5K Dataset.}
Since the core concern of our work is the human body, we crop each image based on the bounding boxes predicted by a person detection algorithm\cite{bochkovskiy2020yolov4} to minimize the impact of backgrounds, which is especially necessary for photos with small persons in a large scene. After the preprocessing, the dataset is randomly divided into a training set with 4,500 images and a testing set with 500 images. In addition, for each image in the training set, we generate its skeleton maps and PAFs in advance using pretrained \emph{Pose Estimator}\cite{cao2017realtime} to speed up the training procedure.
 
\noindent {\bf Implementation Details.} We predefine $\emph{N}_{S}$=12 skeletons and $\emph{N}_{L}$=10 PAFs to represent the body structure, thus the skeleton maps \textbf{S} and PAFs \textbf{L} have 12 and 10 channels respectively. The \emph{Pose Estimator} module is pretrained on COCO keypoint detection dataset and not take part in loss backpropagation. During training, we concatenate the skeleton maps with RGB images to form a stack of input tensors, and augment the input tensors and PAFs identically using random flipping, rotation, and cropping. For computational efficiency, the augmented inputs are resized to 256$\times$256 pixels and then fed into the Flow Generator. The $\theta_g$, $\theta_q$, $\theta_k$ and $\theta_v$ are set as 1$\times$1 convolution in \emph{SASA} module. The weights in loss function are $\lambda_{img}=15$, $\lambda_{flow}=15$ , $\lambda_{orth}=2$. We train our network using Adam optimizer with a learning rate of 2e-5 and batch size of 32.

\noindent {\bf Evaluation Metrics.} We use SSIM, PSNR, and LPIPS\cite{zhang2018unreasonable} to quantitatively evaluate the difference between the reshaped images and the ground-truth images, of which SSIM and PSNR are commonly applied in pixel-level image similarity evaluation, and LPIPS calculates the perceptual similarity in feature space.
 
\subsection{Qualitative Evaluation}
Since 3D model-based body reshaping approaches usually need a dozen minutes of user assistance\cite{zhou2010parametric} or extra sensors\cite{richter2012real,xu2019human}, which heavily hinders their applications in practice, we mainly conduct our comparison with automatic 2D methods. We evaluate four state-of-the-art methods: FAL\cite{wang2019detecting}, ATW\cite{yi2020animating}, pix2pixHD\cite{wang2018high}, and GFLA\cite{ren2020deep}. FAL is designed to detect and reverse the facial warping manipulations, and ATW aims to generate high-quality facial expression animation by motion field prediction, pix2pixHD and GFLA are generative models for image and pose transformation respectively. We retrained these methods on our BR-5K dataset and evaluate their applicability on body reshaping task.

The visual comparison results are shown in Figure~\ref{fig: cmp}. Since FAL and ATW predict flows simply based on RGB information, they can hardly produce satisfied retouching results due to the complexity of human body structure. pix2pixHD is capable of synthesizing photorealistic translation results on high-resolution, but it is generally suitable for pixel-aligned generation, which keeps the semantic unchanged but translates appearance, other than spatial deformation task. We find pix2pixHD usually brings chromatic aberration and artifacts in non-aligned areas (\emph{e.g.}, the red box in the 1st row, and the blue box in the last row). Taking human pose as guidance enables GFLA to create more reasonable results. However, due to estimating global flow fields at the feature level and synthesizing the target by decoding warped features, GFLA can only generate low-resolution results (256$\times$256). Besides, as illustrated in the second row that the arms (red boxes) are uneven on edges, and the last row that the legs (red boxes) are neglected, all of these methods may suffer from incoherent manipulation because of the unawareness of body structural associations. 

By contrast, our approach can produce high-resolution, consistent, and visually pleasing body editing results. The skeletons and PAFs enable us to accurately predict where are crucial for achieving shapely figures, and the \emph{SASA} helps to keep coherent manipulation among body parts. In addition, generating flow at the pixel level empowers us with more controllability in practical usage.

\subsection{Quantitative Evaluation}
\label{sec: Quantitative Evaluation}
We apply the above methods to reshape images from the testing set and compute the aforementioned metrics between their outputs and manually retouched targets. Since pix2pixHD and GFLA cannot handle 2K-resolution images, we upsample their results using bilinear interpolation for a fair evaluation. The quantitative results are shown in Table~\ref{tab:comp}. The Baseline method here means to directly calculate metrics between the input and the target images. According to the comparison, our approach performs favorably against all the other methods on all metrics, which indicates our model is best at imitating artist body retouching strategies. The image-translation methods (pix2pixHD, GFLA) perform worse than flow-based methods (FAL, ATW, ours) due to their resolution limitation and artifacts. Moreover, based on our observation, the LPIPS is more consistent with human perceptual judgments than SSIM and PSNR.

\begin{table}
  \centering
  \caption{Quantitative comparison and user preference on BR-5K dataset. The metrics are the average of 500 test images. $\uparrow$,$\downarrow$ denote if higher or lower is better respectively.} %
    \vspace{-5pt}
  \small
  \begin{tabular}{lcccc}
    \toprule
    Method & SSIM $\uparrow$ & PSNR $\uparrow$ & LPIPS $\downarrow$ & User Study $\uparrow$ \\
    \midrule
    Baseline & 0.8339  & 24.4916 & 0.0823  & N.A. \\
    FAL & 0.8261 & 24.1841  & 0.0837  & 14.4\% \\
    ATW & 0.8316  & 24.6332 & 0.0805  & 9.8\%  \\
    pix2pixHD & 0.7271  & 21.8381 & 0.2800  & 3.6\%  \\
    GFLA  & 0.6649  & 21.4796 & 0.6136  & 1.8\%  \\
    Ours & \textbf{0.8354}  & \textbf{24.7924} & \textbf{0.0777} & \textbf{70.4\%}  \\
    \bottomrule
    \vspace{-30pt}
  \end{tabular}
  \label{tab:comp}
\end{table}

We further conduct a subjective user study for comprehensive evaluation. We invited 40 participants in our experiments, and each participant was shown 30 questions randomly selected from a question pool containing 100 examples. In each question, we show a source human image followed by five body reshaping results of the above four other methods and ours, where the results are arranged in random order. We ask the participants to select the one with best visual quality. As shown in Table~\ref{tab:comp}, the majority of subjects (70.4\% from all) are in favor of our results, demonstrating that our proposed body reshaping method is more visually appealing to the participants than others.

\subsection{Ablation Study}
\begin{figure}[t]
  \centering
  \resizebox{0.98\linewidth}{!}{
   \includegraphics{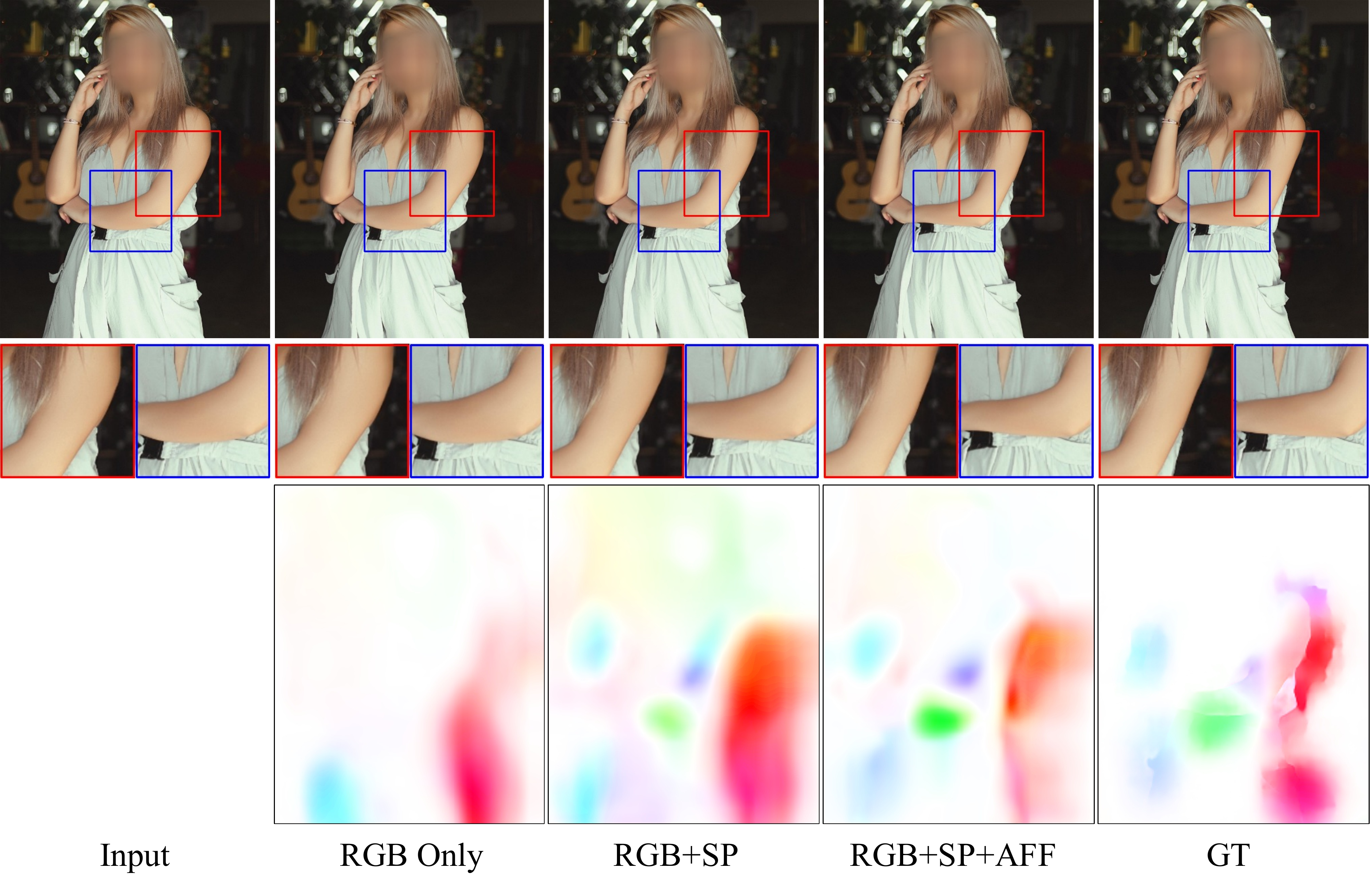} }
       \vspace{-10pt}
  \caption{Integrated with structural priors and the compositional attention, the Full model (RGB+SP+AFF) achieves coherent reshaping result with the succinct but effective deformation flow.}
  \label{fig: ablations}
  \vspace{-10pt}
\end{figure}

To validate the effectiveness of pose priors and \emph{SASA} module, we build two variants of the proposed method by removing structural priors (skeleton maps and PAFs) and \emph{Structure Affinity} block in \emph{SASA} respectively. The model without \emph{Structure Affinity} block is equivalent to a standard self-attention network. Table~\ref{tab:balation} shows their performance. We find that equipped with structural priors and compositional attention, the metrics get steadily improved. As shown in Figure~\ref{fig: ablations}, without the structure guidance, the predicted flow is coarse and vague, resulting in an inadequate reshaping effect. While structural priors enable the flow to focus on arms more precisely, the contour of the upper arm is still uneven, and the background is affected redundantly. By contrast, the Full model generates a succinct but effective flow on both arms and waist areas, leading to a coherent reshaping result while avoiding disturbing background too much, which is more similar to ground-truth than others.
\begin{table}
  \centering
  \caption{Ablation study toward model structures. SP denotes structural priors, AFF denotes \emph{structure affinity} block in \emph{SASA}.}
  \vspace{-10pt}
  \small
  \begin{tabular}{lcccc}
    \toprule
    Method & SSIM $\uparrow$ & PSNR $\uparrow$ & LPIPS $\downarrow$ & EPE $\downarrow$  \\
    \midrule
    w/o SP (RGB only) & 0.8308 &  24.5862  & 0.0789 & 5.0  \\
    w/o AFF (RGB+SP)  & 0.8345 & 24.6449 & 0.0782 & 4.6  \\
    Full (RGB+SP+AFF) & \textbf{0.8354}  & \textbf{24.7924} & \textbf{0.0777} & \textbf{4.1}  \\
    \bottomrule
      \vspace{-15pt}
  \end{tabular}

  \label{tab:balation}
\end{table}

\subsection{Applications}

\noindent {\bf Efficiency on High-Resolution Images.} One advantage of our method is the ability to directly apply the upsampled flow to the original high-resolution image, without introducing discontinuous artifacts. In addition, we test the running speed of our reshaping pipeline (including pose estimation, flow generation, and warping). It takes about 5 seconds to reshape a 4K photo on a 16G Tesla P100 GPU. Consequently, our method provides a feasible option in body retouching workflows with comparable performance to manual results, and boosts the efficiency as well.

\begin{figure}[t]
  \centering
  \resizebox{0.98\linewidth}{!}{
   \includegraphics{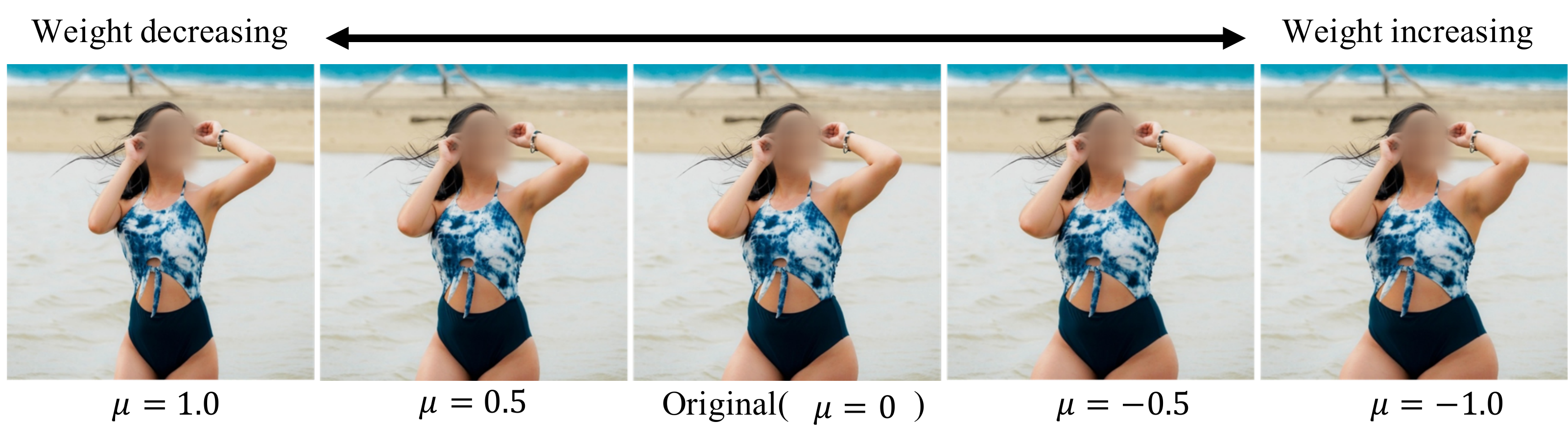} }
       \vspace{-7pt}
  \caption{Our method provides continuous controls on body reshaping by adjusting $\mu$ in the warping procedure.}
  \label{fig: runtime_control}
  \vspace{-5pt}
\end{figure}

\noindent {\bf Runtime Control.} Once the flow field $F_{h}$ is generated, our method can accommodate diverse user preferences by providing continuous controls on the reshaping effects. It can be achieved by simply redoing the warping step with a different multiplier $\mu$ (Equ. \ref{eq: warp_h}). As present in Figure~\ref{fig: runtime_control}, the $|\mu|$ can be set from -1 to 1. A larger $|\mu|$ will introduce a bigger change on body shape, while the positive/negative sign of $\mu$ controls whether to lose or gain weight.

\subsection{Limitation and Discussion}

\begin{figure}[t]
  \centering
  \resizebox{0.98\linewidth}{!}{
   \includegraphics{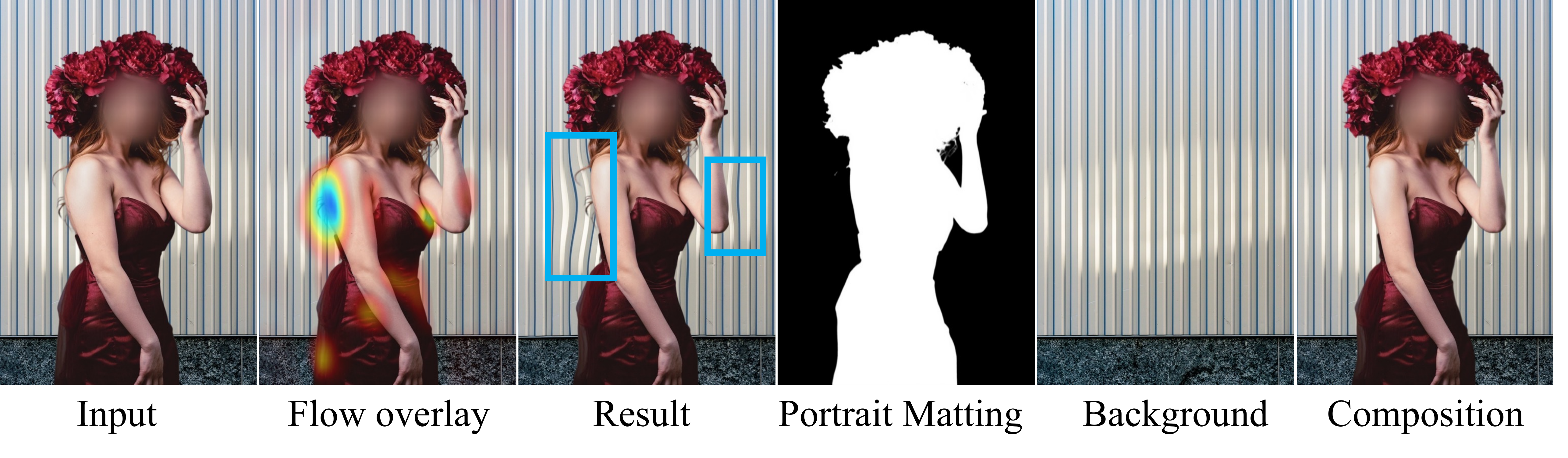} }
       \vspace{-10pt}
  \caption{Flow-based reshaping may bring distortions on the background. One possible solution is to conduct portrait matting and blend the foreground into an identical intact background(either captured in advance or recovered by inpainting).}
  \label{fig: limitation}
  \vspace{-15pt}
\end{figure}

Although our method can achieve stable and visually pleasing body reshaping results, it still has two limitations which we want to discuss. First, As shown in Figure~\ref{fig: limitation}, the predicted flow will also affect the overlapped areas in the background, which may bring some unexpected distortions, such as twisted lines or deformed objects. One possible solution is to combine background matting technique~\cite{sengupta2020background}, segment the reshaped foreground human body and blend it into an identical and intact background image (shown in the last three columns in Figure~\ref{fig: limitation}). However, it should be noted that background matting requires an intact background image which is hard to obtain, and predicting accurate matting results on high-resolution images is difficult. More elegant solutions are to leave in our future work.

Second, a comprehensive body reshaping task involves changing the multidimensional full-body attributes, including both weight and height. However, our proposed approach concentrates on weight editing only, without changing the direction and length of body skeletons. In the meanwhile, reshaping body height can be easily achieved by non-uniform image scaling on body length direction. 

Considering the misuse of the technology may lead to ethical problems, we provide more discussions on the ethical concerns and solutions in the supplemental material.

\section{Conclusion}
In this paper, we propose an end-to-end structure-aware flow generation framework for human body reshaping, which can achieve favorable and controllable results for high-resolution images efficiently. We demonstrate that using the skeletons and PAFs as structural priors can significantly reduce the difficulty for automatic body editing and lead to more accurate deformation flow prediction. A compositional attention mechanism \emph{SASA} is present to improve the manipulation consistency across human parts by considering both perceptual correlations and structural associations. Owing to generating flow in the pixel domain, our method can efficiently handle high-resolution input and support runtime reshaping control. Comprehensive experiments on our proposed BR-5K dataset demonstrate the effectiveness of our method in terms of visual performance, controllability, and efficiency.
 

\clearpage

{\small
\bibliographystyle{ieee_fullname}
\bibliography{egbib}
}

\end{document}